\documentclass[wcp]{jmlr}

\usepackage{booktabs}

\usepackage{float}
\usepackage{graphicx}
\usepackage{array}

\makeatletter
\let\Ginclude@graphics\@org@Ginclude@graphics 
\makeatother

\jmlrvolume{157}
\jmlryear{2019}
\jmlrworkshop{ACML 2019}

\title[Capsule Network for Anomaly Detection]{Anomaly Detection using Capsule Networks for High-dimensional Datasets}

 \author{\Name{Inderjeet Singh} \Email{inderjeet78@iitb.ac.in}\and
  \Name{Nandyala Hemachandra} \Email{nh@iitb.ac.in}\\
  \addr Indian Institute of Technology Bombay,Mumbai,India}

\editors{Wee Sun Lee and Taiji Suzuki}

\begin{document}

\maketitle

\begin{abstract}
Anomaly detection is an essential problem in machine learning. Application areas include network security, health care, fraud detection, etc., involving high-dimensional datasets. A typical anomaly detection system always faces the class-imbalance problem in the form of a vast difference in the sample sizes of different classes. They usually have class overlap problems. This study used a capsule network for the anomaly detection task. To the best of our knowledge, this is the first instance where a capsule network is analyzed for the anomaly detection task in a high-dimensional complex data setting. We also handle the related novelty and outlier detection problems. The architecture of the capsule network was suitably modified for a binary classification task. Capsule networks offer a good option for detecting anomalies due to the effect of viewpoint invariance captured in its predictions and viewpoint equivariance captured in internal capsule architecture. We used six-layered under-complete autoencoder architecture with second and third layers containing capsules. The capsules were trained using the dynamic routing algorithm. We created $10$-imbalanced datasets from the original MNIST dataset and compared the performance of the capsule network with $5$ baseline models. Our leading test set measures are F1-score for minority class and area under the ROC curve. We found that the capsule network outperformed every other baseline model on the anomaly detection task by using only ten epochs for training and without using any other data level and algorithm level approach. Thus, we conclude that capsule networks are excellent in modeling complex high-dimensional imbalanced datasets for the anomaly detection task.
\end{abstract}

\section{Introduction}
The problem of anomaly detection refers to the problem of finding patterns in data that do not conform to expected behavior, or simply anomaly detection is a method of detecting unusual behavioral patterns in a system. These anomalous behaviors have different names in domains like anomalies, outliers, discordant observations, exceptions, aberrations, surprises or contaminants, etc. Typically the anomalous items translate to some kind of problems such as bank fraud, a structural defect, medical problems, errors in a text, network intrusion, event detection in sensor network, or ecosystem disturbances \cite{sommer2010outside}. Anomalies are also referred as outliers, novelties, noise, deviations, and exceptions. Most of the time, in practical systems, we are interested in rarely occurring instances like network intrusion, defect in a machine in a large factory, credit card fraud, suspicious patterns in a production process in a manufacturing environment, detecting ability to repay the loan of a regular customer by monitoring his/her financial behavior, etc. We are highly interested in those instances because of their huge impact on the entire system in their respective domains. Also, note that novelty detection and outlier detection are two different approaches in anomaly detection. The difference between them can be seen from below:\\
\textbf{Novelty Detection:}
When training data is not polluted by outliers, and we are interested in detecting anomalies in the new observations. This concept is specially used by unsupervised algorithms such as autoencoders and one-class SVM (~\cite{ocsvm}).\\
\textbf{Outlier Detection:}
When training data contains outliers, and we need a central tendency measure of the training data, ignoring the deviant observations. This concept is mainly used by all the supervised learning algorithms and unsupervised learning algorithms such as Isolation Forest (~\cite{liu2008isolation}), Local outlier factor (~\cite{breunig2000lof}, DBSCAN (~\cite{ester1996density}), etc.
\par Basic requirement of an excellent anomaly detection system is that it must perform well for both novelty detection as well as outlier detection.

\subsection*{Anomaly Detection Techniques}
Anomaly detection techniques can be broadly categorized into three categories: supervised learning methods, unsupervised learning methods, and semi-supervised learning methods.
Some of the popular anomaly detection techniques used by now are distance-based techniques, density-based techniques (used by algorithms such as k-nearest neighbor, local outlier factor, isolation forest, etc.), Ensemble techniques, one-class learning methods, Bayesian networks, hidden Markov models, cluster-based outlier detection, fuzzy logic based outlier detection, Replicator neural networks, Autoencoders, Deviation from association rules and frequent itemsets, Subspace \& correlation-based outlier detection.

The importance of anomaly detection is due to the fact that most of the time, anomalous instances in a system translate to the most critical information in a system. Cybercrime and threat activities have become a critical part of our daily life, and the importance of network security has continuously emerged as a significant concern ~\cite{kwon2017survey}. 
There is a wide range of domains in which anomaly detection plays a critical role.
We can take the example of detection of abnormal behavior in crowded scenes where ~\cite{sabokrou2018deep} used Fully Convolutional Neural Network(FCN) on temporal data. They transferred pre-trained FCN to unsupervised FCN to ensure the detection of global anomalies. ~\cite{du2017deeplog} proposed DeepLog, a deep neural network utilizing Long Short-Term Memory (LSTM) to model system log as a natural language sequence. DeepLog records system states and significant events at various critical points to help debug system failures and perform root cause analysis. ~\cite{lu2018motor} developed an anomaly detection system using reinforcement learning to prevent the motor of the drone from operating at abnormal temperatures. The proposed system automatically lands the drone once the temperature exceeds the threshold. Anomaly detection is also used in malware detection in computer programs, ~\cite{kharche2015internet} proposed a general worm detection model where they used Bayesian Network, decision tree, and random forest classifiers in their model.
\par Nowadays, deep learning models are getting more and more popular in anomaly detection tasks due to their superior performance compared to other classical machine learning models due to their better ability to model complex high-dimensional data. ~\cite{an2015variational} proposed variational autoencoder-based anomaly detection using reconstruction probability which outperformed earlier autoencoder and principle components based anomaly detection methods. ~\cite{zhou2017anomaly} proposed a more Robust Deep Autoencoder(RDA) architecture for eliminating noise and outliers without needing to access the clean data. Their model was inspired by Robust Principal Component Analysis, where they split the data $X$ into two parts, $X = L_D + S$, where $L_D$ is effectively reconstructed by deep autoencoder $S$ contains the outliers and the noise in the original data $X$. This method showed significant improvement of $73\%$ over Isolation Forests.\cite{deecke2018image} showed that searching the latent space of the generator of Generative Adversarial Network(GAN) can be leveraged for use in anomaly detection tasks. They achieved state-of-the-art performance on standard image benchmark datasets. Their model can be used for the unsupervised anomaly detection task.~\cite{schlegl2017unsupervised} used GANs for unsupervised anomaly detection to guide marker discovery. ~\cite{zong2018deep} proposed a deep autoencoding Gaussian mixture model for unsupervised anomaly detection. They outperformed state-of-the-art anomaly detection techniques on high-dimensional datasets. ~\cite{zenati2018efficient} used efficient GAN-based anomaly detection methods and achieved state-of-the-art performance while being several hundred-fold faster at test time than the recently published GAN-based method back then. ~\cite{ngo2019fence} proposed a new GAN architecture called fence GAN (FGAN) with modified GAN loss. FGAN  directly uses the discriminator score as an anomaly threshold. ~\cite{li2019mad} did multivariate anomaly detection on time series data using GANs and named their model as MAD-GAN.

\subsection*{Class Overlap Problem in Anomaly Detection}
When the anomalous instances occur in very dense sub-spaces occupied mostly by standard instances, then in these scenarios, machine learning algorithms cannot detect those minority classes or require a highly complex algorithm to detect them. This problem is called the class overlap problem in machine learning \cite{bellinger2017sampling}. In simple words, when a small cluster of minority class instances lies inside a dense cluster of majority class instances in latent spaces, the poor performance of learning algorithms on minority classes is due to the class overlap problem. Most of the time, this problem occurs when the dataset is missing some essential features or due to improper handling of data. This problem is common in non-image datasets.

\section{Capsule Networks}
Geoffrey Hinton introduced the idea of capsules in a neural network, then ~\cite{sabour2017dynamic} introduced an entirely new type of neural network based on capsules. They provided six-layered autoencoder-based architecture having a capsule layer in the encoder part.
They published an algorithm called \textit{dynamic routing between capsules} that allowed to train such a network. The main characteristic of this network is the viewpoint invariance for the detection of classes due to viewpoint equivariance captured by internal capsule architecture. It allows this network to correctly label unobserved patterns with very high probability. This network is more robust for unobserved anomalous and normal instances.
Later they provided improved architecture of capsule net (\cite{hinton2018matrix}) where they used a pose matrix for each capsule instead of a vector in the earlier model. For the routing of capsules in the network, they used Expectation-Maximization(EM) algorithm. On the smallNORB benchmark dataset, this model reduced the test errors by $45\%$ compared to the state-of-the-art. They also showed that the capsule network is robust against the white box adversarial attacks, which seems to be promising for their application in anomaly detection problems.

\subsection{Why Capsule Networks for Anomaly Detection}
Capsule Network was initially proposed to better mimic the biological neural organization by overcoming the drawbacks present in Convolutional Neural Networks(CNN). In CNNs, pooling allows a degree of translational invariance and a large number of feature types to be represented. Capsule theory argues that pooling in CNNs violates biological shape perception and ignores the underlying linear manifold in images. Pooling also causes loss of important information by throwing out data, thus damaging nearby feature detectors. The biggest advantage of capsule networks is that it provides equivariance in internal architecture (trained capsules) while, on the other hand, CNNs provide invariance due to discarding positional information.  

The aim of capsules is to use underlying linearity to deal with viewpoint variations and improve segmentation decisions. Capsules use high dimensional coincidence filtering. ~\cite{zhou2016spatial} used spatial-temporal CNNs for anomaly detection and localization in crowded scenes and outperformed state-of-the-art methods. Considering the advantages of capsule networks over CNNs, they appear to be excellent candidates for the anomaly detection task. We know that anomalies are few and different from normal instances in a system. Most of the time, newer anomalies follow completely new unobserved patterns occurring in a system. Since ~\cite{sabour2017dynamic} already shown that capsule networks have excellent viewpoint invariance properties for detecting the class of an instance, and also they have viewpoint equivariance property in the internal architecture due to the presence of the capsules, they can model unobserved patterns better as compared to CNNs and other present state-of-the-art methods. Another advantage of capsules networks is that they require fewer data to achieve the same performance as other deep learning models. So if we are not performing unsupervised anomaly detection, then capsule networks are a good choice due to the scarcity of anomalous instances in the training set. Also, using them as supervised learning methods and having their considerably better ability for classifying highly overlapped instances(~\cite{sabour2017dynamic}), they solve the class overlap problem which unsupervised learning models could not solve. 

The better ability of capsule networks for the class overlap is that the parallel attention mechanism of this network allows each capsule at one level to attend to some active capsules at the level below and ignore others. So when multiple objects are present in an instance, active lower level capsules send their outputs to their respective parent capsules. So multiple parent capsules become active, and the process goes on till the output layer has multiple capsules indicating the presence of a class. So due to all these facts, we believe that capsule networks are an excellent choice for anomaly detection.

\section{Capsule Network Architecture for anomaly detection}
We used six-layered autoencoder architecture containing two layers of capsules similar to the one proposed in ~\cite{sabour2017dynamic}. The network has two parts, an encoder and a decoder having three layers each. The network's input layer takes $28$x$28$ pixel images of the MNIST dataset. The first hidden layer is a convolution layer that outputs a tensor of size $20$x$20$x$256$ using ReLu activation and without padding. 

\begin{figure}
    \centering
    \includegraphics[scale=0.4]{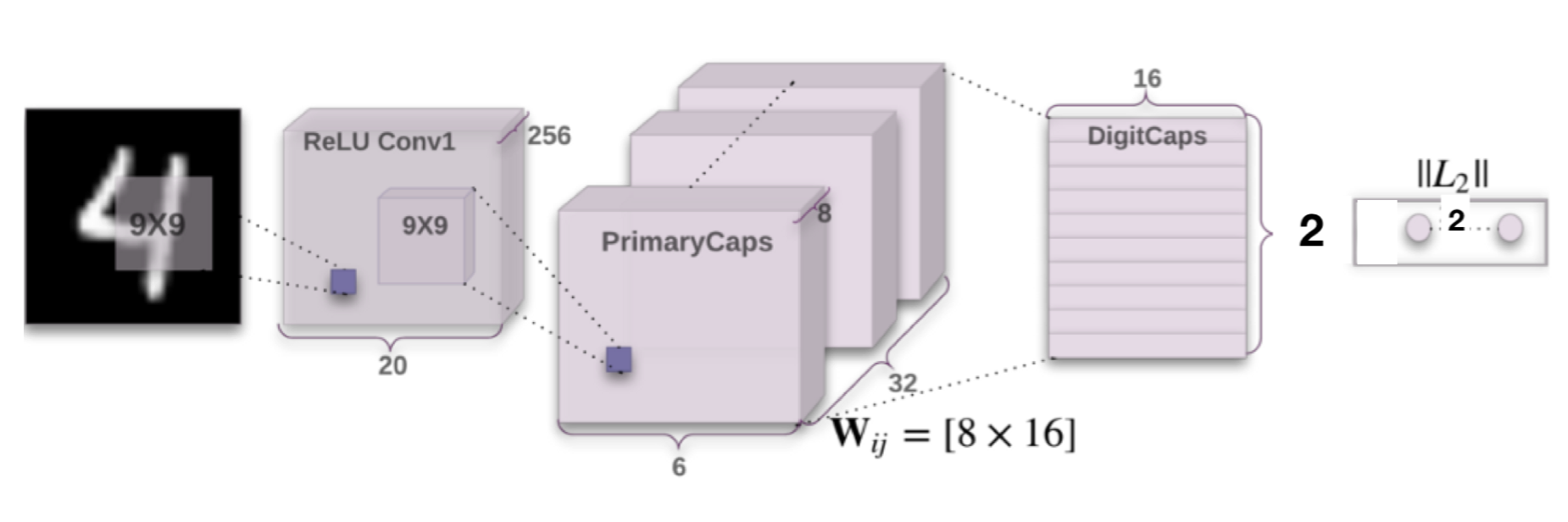}
    \caption{Encoder part of the capsule network}
    \label{fig:capsencoder}
\end{figure}

The second hidden layer was the first layer containing capsules, outputs a tensor of size $6$x$6$x$8$x$32$. The output of the second layer can be viewed as $6$x$6$x$32$ 8-dimensional capsules. The third layer of the network is also the output layer of the encoder, and the second capsule layer of the network, outputs two 16-dimensional vectors, one for each class. Here at the place of two capsules, we could have taken one capsule only because the length of the capsule represents the presence of a class, and we are dealing with binary classification here. Nevertheless, as per capsule theory, each capsule's instantiation parameters should be recording one class because allowing a capsule to record more than one class's instantiation parameters may induce high bias in the network. Due to this, we kept a second capsule for recording anomalous instances. However, the size of the second capsule depends upon the imbalance and nature of anomalies in the data, which is a part of another discussion. The decoder part of the network has three fully connected layers outputting a number of neurons as $512$,$1024$, and $784$, respectively. The output of the decoder is then reshaped back to size $28$x$28$ to see the reconstructed image.
\par The loss function of the encoder of this network is very similar to the SVM loss function. Here is the loss function for the encoder part:
$$L_c = T_c(max(0,m^+ - ||v_c||)^2) + \lambda (1-T_c)(max(0,||v_c|| - m^-)^2)$$
Here,
$L_c$ is loss term for one DigitCap.
$v_c$ is DigitCap vector.
$T_c$ is 1 for correct digitcap and 0 for incorrect digitcap, basically this represent true label of an instance.
$\lambda$ is a constant they originally used for numerical stability.
$m^+$ and $m^-$ are constants which are fixed at 0.9 and 0.1 respectively.
The loss function for decoder (reconstruction loss) is simply the mean squared error let it denote by $L_d$.
So the loss function for entire network is as follows:\\
$$L = \sum_c L_c + \alpha. L_d$$
So in the total loss, reconstruction loss acts as a regularizer whose strength is controlled using parameter $\alpha$.

\section{Experiments and discussion}
\subsection{Data description}
The experiments were performed on standard MNIST data (~\cite{mnist}). The training set of this data contains $60000$, $28$x$28$ pixel images of handwritten digits from $0$ to $9$. The test set of this data contains $10000$ images of the same size. A validation set of $5000$ images was prepared using stratified random sampling from training data. Inherently this is not highly imbalanced data. Since our objective is to assess the performance of capsule network for anomaly detection task, we have made this data imbalanced and converted it to binary classification problem from multi-class classification problem. We created $10$ subsets for each training, validation, and test dataset. These $10$ subsets correspond to each of the $10$ classes. For example, for training subset-$0$, we took all the samples belonging to class $0$ from training data, randomly sampled instances from other classes from the same training data, and labeled them all as $1$. For each subset, we labeled the majority class or normal class as $0$ and the class of anomalous instances as $1$. The number of sampled instances from other classes (anomalous instances)  was kept around $1/10^{th}$ of the number of instances of class $0$.

Similarly, validation subset-$0$ and test subset-$0$ were created. So there were $10$ training subsets, $10$ validation, and $10$ test subset corresponding to each class where the subset number indicates which class is in the majority in the dataset. So we created $10$ binary class imbalanced datasets from MNIST data to evaluate the performance of our capsule network on anomaly detection task.

\subsection{Baseline Methods}
We considered traditional, popular, and state-of-the-art anomaly detection methods.
\begin{itemize}
    \item \textbf{Random Forest Classifier:} This is a popular ensemble learning decision tree-based method proposed by ~\cite{Breiman2001}. This method performs very well most of the time for supervised learning tasks. We tuned its important parameters during training, like the number of trees in the forest and the minimum number of samples required at a node for the further split. We also enabled cost-sensitive learning by tuning class weight parameters.
    \item \textbf{XGBoost Classifier:} XGBoost is a tool proposed by ~\cite{friedman2001greedy} and motivated by formal principles of machine learning, mathematical optimization, and computational efficiency of learning algorithms. Due to its recent success in many machine learning-based competitions held on Kaggle and other platforms, we also decided to include this model for comparative analysis. We kept the base classifier of this model as gradient boosted trees and tuned the learning rate, number of base estimators, minimum child weight, and $\gamma$ hyperparameters. 
    \item \textbf{Deep Neural Network:} We used six layers deep neural network architecture with \textit{relu} activation function for hidden layers and sigmoid activation function for the output layer. The number of hidden units in each layer was also tuned along with activation functions.
    \item \textbf{Isolation Forest Classifier:} This is an unsupervised learning method for outlier detection proposed by ~\cite{liu2008isolation}. This is one of the newest techniques to detect anomalies. This algorithm is based on the isolation mechanism according to which anomalies can be isolated much earlier than a normal instance due to their very different nature than normal instances. We tuned a number of the base estimator and maximum features to draw from training set hyper-parameters.
    \item \textbf{Undercomplete Deep Autoencoder:} We used six-layered undercomplete deep autoencoder architecture, three layers each for encoder and decoder part. We used the concept of reconstruction probability for binary classification in our case. Hidden layer sizes and activation functions were fully tuned. For classification, we used reconstruction probability similar to ~\cite{an2015variational}.
\end{itemize}
\begin{figure}
    \centering
    \includegraphics[scale=0.48]{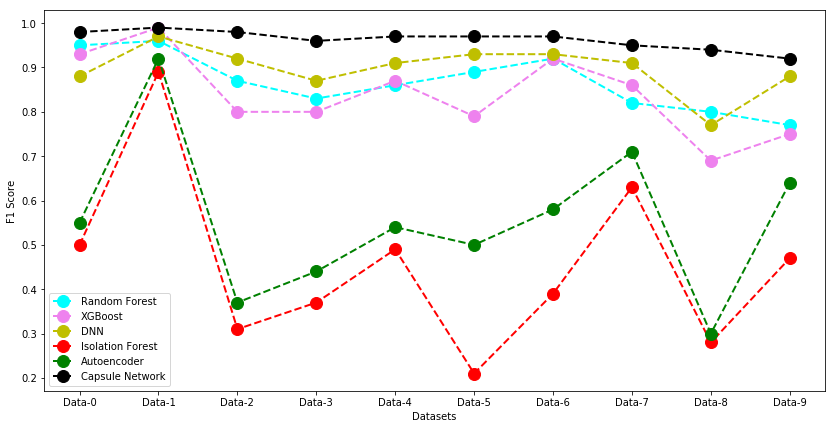}
    \caption{F1-Score of Capsule Network and baseline models on test datasets for the minority class}
    \label{fig:f1}
\end{figure}
\subsection{Results}
We evaluated the performance of capsule network and baseline methods on all the $10$ imbalanced datasets created from the MNIST dataset as described earlier. We kept the F1-score for minority class as the main evaluation metric and Area Under ROC (AUROC) and overall accuracy. For the \textbf{Dataset-0}, we observed the following results:
\begin{table}
\centering
\begin{center}
\begin{tabular}{ |p{3.6cm}|p{1.55cm}|p{1.55cm}|p{1.55cm}|p{1.55cm}|p{1.55cm}|}
\hline
\centering Models &  Accuracy & AUROC &  Precision &  Recall  &  F1-score  \\
\hline
Random Forest   & 98.84  \%& 0.988   &   0.96   & 0.94 & 0.95  \\\hline
XGBoost         & 98.30  \%&  0.994  &   0.97   & 0.89 & 0.93  \\\hline
DNN             & 97.22  \%&   0.996 &   0.97   & 0.80 & 0.88  \\\hline
Isolation Forest& 97.62  \%&  0.680  &   0.73   & 0.38 & 0.50  \\\hline
Autoencoder     & 85.67  \%&  0.892  &   0.45   & 0.72 & 0.55  \\\hline
Capsule Network & \textbf{99.47}  \%& \textbf{ 0.999}  &   \textbf{1.00}   &\textbf{ 0.96 }& \textbf{0.98}  \\
\hline
\end{tabular}
\end{center}
\caption{The test set results for Dataset-0 (Precision, recall, and f1-score are for minority class)}
\label{tab:d0}
\end{table}

Clearly, in table \ref{tab:d0}, we can observe that on dataset-0, the capsule network is outperforming every other baseline model. The same can also be observed from figure \ref{fig:f1} given below.


Apart from results on dataset-0, we saw that the capsule network outperformed every other baseline model on all the remaining datasets (from dataset-1 to dataset-9), which can be seen from Appendix A and Appendix B.
\begin{figure}
    \centering
    \includegraphics[scale=0.48]{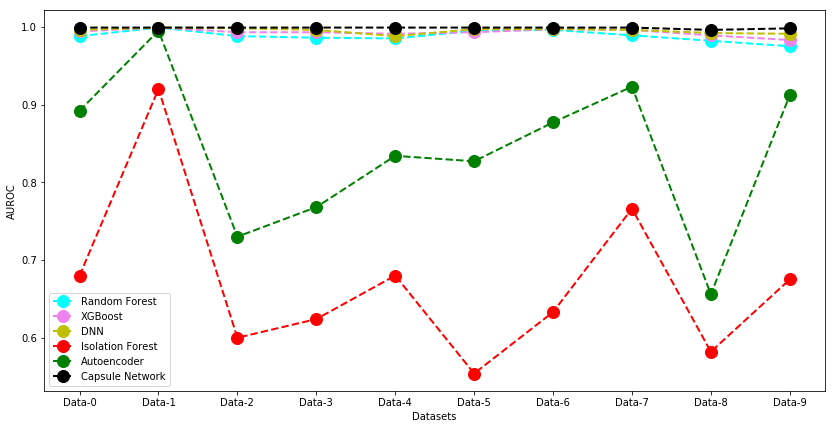}
    \caption{Area Under ROC for Capsule Network and baseline models on  test datasets}
    \label{fig:auroc}
\end{figure}
Capsule network is shown superior results for all evaluation metrics used here, which are F1-score, AUROC, and accuracy. Since the data is high-dimensional, reasonably complex, and imbalanced, we can say that the capsules network could model this data really well, which supports our arguments made in section $2.1$. So we have observed that capsules networks have a high potential to model complex high-dimensional imbalanced data compared to other popular anomaly detection models. In the present study, we used dynamic routing-based capsules in our network, which is an encouraging sign to go for expectation-maximization-based capsules networks (~\cite{hinton2018matrix}). 

\section{Conclusion}
Anomaly detection is the problem of detecting unexpected, unusual behavioral patterns in a system. In this domain, we deal with highly imbalanced datasets. We have seen that supervised machine learning algorithms like Random forest, XGBoost, deep neural networks, and unsupervised learning models like isolation forest, one-class SVM, and autoencoders perform well for low dimensional data. We also use data level approaches like undersampling and oversampling, algorithm level approaches like cost-sensitive learning, and different calibration techniques to improve the results. However, these algorithms struggle most of the time on high-dimensional complex data. 
\par ~\cite{sabour2017dynamic} and ~\cite{hinton2018matrix} shown that capsule networks are excellent in modeling viewpoint invariance in predictions due to viewpoint equivariance captured by internal capsule architecture. This encouraged us to use capsule networks for anomaly detection tasks where viewpoint invariance and viewpoint equivariance play an essential role. We conducted our results on imbalanced MNIST datasets created from original MNIST data (~\cite{mnist}). We used random forest, XGBoost, deep neural network, isolation forest, and undercomplete deep autoencoder as our baseline models to compare the performance of capsule network on this data. We observed that for training up to 10 epoch only, the capsules network outperformed every other baseline model on all high-dimensional datasets we created from MNIST data. We can conclude that capsule networks are excellent in modeling complex high-dimensional imbalanced datasets from these results.


\bibliography{acml21}

\begin{thebibliography}{25}
\providecommand{\natexlab}[1]{#1}
\providecommand{\url}[1]{\texttt{#1}}
\expandafter\ifx\csname urlstyle\endcsname\relax
  \providecommand{\doi}[1]{doi: #1}\else
  \providecommand{\doi}{doi: \begingroup \urlstyle{rm}\Url}\fi

\bibitem[An and Cho(2015)]{an2015variational}
Jinwon An and Sungzoon Cho.
\newblock Variational autoencoder based anomaly detection using reconstruction
  probability.
\newblock \emph{Special Lecture on IE}, 2:\penalty0 1--18, 2015.

\bibitem[Bellinger et~al.(2017)Bellinger, Sharma, Za{\i}ane, and
  Japkowicz]{bellinger2017sampling}
Colin Bellinger, Shiven Sharma, Osmar~R Za{\i}ane, and Nathalie Japkowicz.
\newblock Sampling a longer life: Binary versus one-class classification
  revisited.
\newblock In \emph{First International Workshop on Learning with Imbalanced
  Domains: Theory and Applications}, pages 64--78, 2017.

\bibitem[Breiman(2001)]{Breiman2001}
Leo Breiman.
\newblock {Random forest}.
\newblock \emph{Machine Learning}, 2001.
\newblock ISSN 18790534.
\newblock \doi{10.1016/j.compbiomed.2011.03.001}.

\bibitem[Breunig et~al.(2000)Breunig, Kriegel, Ng, and Sander]{breunig2000lof}
Markus~M Breunig, Hans-Peter Kriegel, Raymond~T Ng, and J{\"o}rg Sander.
\newblock Lof: identifying density-based local outliers.
\newblock In \emph{ACM sigmod record}, volume~29, pages 93--104. ACM, 2000.

\bibitem[Deecke et~al.(2018)Deecke, Vandermeulen, Ruff, Mandt, and
  Kloft]{deecke2018image}
Lucas Deecke, Robert Vandermeulen, Lukas Ruff, Stephan Mandt, and Marius Kloft.
\newblock Image anomaly detection with generative adversarial networks.
\newblock In \emph{Joint European Conference on Machine Learning and Knowledge
  Discovery in Databases}, pages 3--17. Springer, 2018.

\bibitem[Du et~al.(2017)Du, Li, Zheng, and Srikumar]{du2017deeplog}
Min Du, Feifei Li, Guineng Zheng, and Vivek Srikumar.
\newblock Deeplog: Anomaly detection and diagnosis from system logs through
  deep learning.
\newblock In \emph{Proceedings of the 2017 ACM SIGSAC Conference on Computer
  and Communications Security}, pages 1285--1298. ACM, 2017.

\bibitem[Ester et~al.(1996)Ester, Kriegel, Sander, Xu,
  et~al.]{ester1996density}
Martin Ester, Hans-Peter Kriegel, J{\"o}rg Sander, Xiaowei Xu, et~al.
\newblock A density-based algorithm for discovering clusters in large spatial
  databases with noise.
\newblock In \emph{Kdd}, volume~96, pages 226--231, 1996.

\bibitem[Friedman(2001)]{friedman2001greedy}
Jerome~H Friedman.
\newblock Greedy function approximation: a gradient boosting machine.
\newblock \emph{Annals of statistics}, pages 1189--1232, 2001.

\bibitem[Hinton et~al.(2018)Hinton, Sabour, and Frosst]{hinton2018matrix}
Geoffrey~E Hinton, Sara Sabour, and Nicholas Frosst.
\newblock Matrix capsules with em routing.
\newblock 2018.

\bibitem[Kharche and Thakare(2015)]{kharche2015internet}
Dipali Kharche and Anuradha Thakare.
\newblock Internet worm classification and detection using data mining
  techniques, 2015.

\bibitem[Kwon et~al.(2017)Kwon, Kim, Kim, Suh, Kim, and Kim]{kwon2017survey}
Donghwoon Kwon, Hyunjoo Kim, Jinoh Kim, Sang~C Suh, Ikkyun Kim, and Kuinam~J
  Kim.
\newblock A survey of deep learning-based network anomaly detection.
\newblock \emph{Cluster Computing}, pages 1--13, 2017.

\bibitem[LeCun and Cortes(2010)]{mnist}
Yann LeCun and Corinna Cortes.
\newblock {MNIST} handwritten digit database.
\newblock 2010.
\newblock URL \url{http://yann.lecun.com/exdb/mnist/}.

\bibitem[Li et~al.(2019)Li, Chen, Shi, Jin, Goh, and Ng]{li2019mad}
Dan Li, Dacheng Chen, Lei Shi, Baihong Jin, Jonathan Goh, and See-Kiong Ng.
\newblock Mad-gan: Multivariate anomaly detection for time series data with
  generative adversarial networks.
\newblock \emph{arXiv preprint arXiv:1901.04997}, 2019.

\bibitem[Liu et~al.(2008)Liu, Ting, and Zhou]{liu2008isolation}
Fei~Tony Liu, Kai~Ming Ting, and Zhi-Hua Zhou.
\newblock Isolation forest.
\newblock In \emph{2008 Eighth IEEE International Conference on Data Mining},
  pages 413--422. IEEE, 2008.

\bibitem[Lu et~al.(2018)Lu, Li, Mu, Wang, Kim, and Serikawa]{lu2018motor}
Huimin Lu, Yujie Li, Shenglin Mu, Dong Wang, Hyoungseop Kim, and Seiichi
  Serikawa.
\newblock Motor anomaly detection for unmanned aerial vehicles using
  reinforcement learning.
\newblock \emph{IEEE internet of things journal}, 5\penalty0 (4):\penalty0
  2315--2322, 2018.

\bibitem[Ngo et~al.(2019)Ngo, Winarto, Li, Park, Akram, and Lee]{ngo2019fence}
Cuong~Phuc Ngo, Amadeus~Aristo Winarto, Connie Kou~Khor Li, Sojeong Park,
  Farhan Akram, and Hwee~Kuan Lee.
\newblock Fence gan: Towards better anomaly detection.
\newblock \emph{arXiv preprint arXiv:1904.01209}, 2019.

\bibitem[Sabokrou et~al.(2018)Sabokrou, Fayyaz, Fathy, Moayed, and
  Klette]{sabokrou2018deep}
Mohammad Sabokrou, Mohsen Fayyaz, Mahmood Fathy, Zahra Moayed, and Reinhard
  Klette.
\newblock Deep-anomaly: Fully convolutional neural network for fast anomaly
  detection in crowded scenes.
\newblock \emph{Computer Vision and Image Understanding}, 172:\penalty0 88--97,
  2018.

\bibitem[Sabour et~al.(2017)Sabour, Frosst, and Hinton]{sabour2017dynamic}
Sara Sabour, Nicholas Frosst, and Geoffrey~E Hinton.
\newblock Dynamic routing between capsules.
\newblock In \emph{Advances in neural information processing systems}, pages
  3856--3866, 2017.

\bibitem[Schlegl et~al.(2017)Schlegl, Seeb{\"o}ck, Waldstein, Schmidt-Erfurth,
  and Langs]{schlegl2017unsupervised}
Thomas Schlegl, Philipp Seeb{\"o}ck, Sebastian~M Waldstein, Ursula
  Schmidt-Erfurth, and Georg Langs.
\newblock Unsupervised anomaly detection with generative adversarial networks
  to guide marker discovery.
\newblock In \emph{International Conference on Information Processing in
  Medical Imaging}, pages 146--157. Springer, 2017.

\bibitem[Sch{\"o}lkopf et~al.(2001)Sch{\"o}lkopf, Platt, Shawe-Taylor, Smola,
  and Williamson]{ocsvm}
Bernhard Sch{\"o}lkopf, John~C Platt, John Shawe-Taylor, Alex~J Smola, and
  Robert~C Williamson.
\newblock Estimating the support of a high-dimensional distribution.
\newblock \emph{Neural computation}, 13\penalty0 (7):\penalty0 1443--1471,
  2001.

\bibitem[Sommer and Paxson(2010)]{sommer2010outside}
Robin Sommer and Vern Paxson.
\newblock Outside the closed world: On using machine learning for network
  intrusion detection.
\newblock In \emph{Security and Privacy (SP), 2010 IEEE Symposium on}, pages
  305--316. IEEE, 2010.

\bibitem[Zenati et~al.(2018)Zenati, Foo, Lecouat, Manek, and
  Chandrasekhar]{zenati2018efficient}
Houssam Zenati, Chuan~Sheng Foo, Bruno Lecouat, Gaurav Manek, and
  Vijay~Ramaseshan Chandrasekhar.
\newblock Efficient gan-based anomaly detection.
\newblock \emph{arXiv preprint arXiv:1802.06222}, 2018.

\bibitem[Zhou and Paffenroth(2017)]{zhou2017anomaly}
Chong Zhou and Randy~C Paffenroth.
\newblock Anomaly detection with robust deep autoencoders.
\newblock In \emph{Proceedings of the 23rd ACM SIGKDD International Conference
  on Knowledge Discovery and Data Mining}, pages 665--674. ACM, 2017.

\bibitem[Zhou et~al.(2016)Zhou, Shen, Zeng, Fang, Wei, and
  Zhang]{zhou2016spatial}
Shifu Zhou, Wei Shen, Dan Zeng, Mei Fang, Yuanwang Wei, and Zhijiang Zhang.
\newblock Spatial--temporal convolutional neural networks for anomaly detection
  and localization in crowded scenes.
\newblock \emph{Signal Processing: Image Communication}, 47:\penalty0 358--368,
  2016.

\bibitem[Zong et~al.(2018)Zong, Song, Min, Cheng, Lumezanu, Cho, and
  Chen]{zong2018deep}
Bo~Zong, Qi~Song, Martin~Renqiang Min, Wei Cheng, Cristian Lumezanu, Daeki Cho,
  and Haifeng Chen.
\newblock Deep autoencoding gaussian mixture model for unsupervised anomaly
  detection.
\newblock 2018.

\end{thebibliography}

\appendix

\section{First Appendix}\label{apd:first}

Experimental results on MNIST dataset. The values of precision, recall, and f1-score are for minority class which is of our interest. We kept Accuracy and AUROC results to get the overall performance of classifiers for both classes.

\begin{table}
\centering
\begin{center}
\begin{tabular}{ |p{3.6cm}|p{1.55cm}|p{1.55cm}|p{1.55cm}|p{1.55cm}|p{1.55cm}|}
\hline
\centering Models &  Accuracy & AUROC & Precision & Recall  & F1-score  \\
\hline
Random Forest   & 99.06  \%& 0.999   &   0.96   & 0.95 & 0.96  \\\hline
XGBoost         & 98.82  \%&  0.999  &   0.99   & 0.99 & 0.99  \\\hline
DNN             & 99.45  \%&  0.999  &   0.99   & 0.96 & 0.97  \\\hline
Isolation Forest& 97.72  \%&  0.920  &   0.94   & 0.85 & 0.89  \\\hline
Autoencoder     & 98.27  \%&  0.995  &   0.88   & 0.97 & 0.92  \\\hline
Capsule Network & \textbf{99.98} \% &  \textbf{0.999}  &   1.00   & 0.98 & \textbf{0.99}  \\
\hline
\end{tabular}
\end{center}
\caption{Test set results for Dataset-1}
\label{tab:st ens}
\end{table}

\begin{table}
\centering
\begin{center}
\begin{tabular}{ |p{3.6cm}|p{1.55cm}|p{1.55cm}|p{1.55cm}|p{1.55cm}|p{1.55cm}|}
\hline
\centering Models &  Accuracy & AUROC & Precision & Recall  & F1-score  \\
\hline
Random Forest   & 96.58  \%&  0.988  &   0.80   & 0.94 & 0.87  \\\hline
XGBoost         & 95.89  \%&  0.993  &   0.94   & 0.69 & 0.80  \\\hline
DNN             & 98.20  \%& 0.998   &   0.95   & 0.89 & 0.92  \\\hline
Isolation Forest& 88.11  \%&  0.60  &   0.48   & 0.23 & 0.31  \\\hline
Autoencoder     & 74.17  \%&  0.730  &   0.26   & 0.64 & 0.37  \\\hline
Capsule Network & \textbf{99.58} \% &   \textbf{0.999} &   0.99   & 0.97 & \textbf{0.98}  \\
\hline
\end{tabular}
\end{center}
\caption{Test set results for Dataset-2}
\label{tab:st ens}
\end{table}

\begin{table}[H]
\centering
\begin{center}
\begin{tabular}{ |p{3.6cm}|p{1.55cm}|p{1.55cm}|p{1.55cm}|p{1.55cm}|p{1.55cm}|}
\hline
\centering Models &  Accuracy & AUROC & Precision & Recall  & F1-score  \\
\hline
Random Forest   & 95.81  \%&  0.986  &   0.81   & 0.85 & 0.83  \\\hline
XGBoost         & 95.89  \%&  0.993  &   0.94   & 0.69 & 0.80  \\\hline
DNN             & 97.21  \%&  0.996  &   0.99   & 0.77 & 0.87  \\\hline
Isolation Forest& 88.75  \%&  0.624  &   0.56   & 0.28 & 0.37  \\\hline
Autoencoder     & 83.78  \%&  0.768  &   0.38   & 0.54 & 0.44  \\\hline
Capsule Network & \textbf{99.13} \% & \textbf{0.999}   &   1.00   & 0.93 & \textbf{0.96} \\
\hline
\end{tabular}
\end{center}
\caption{Test set results for Dataset-3}
\label{tab:st ens}
\end{table}

\begin{table}[H]
\centering
\begin{center}
\begin{tabular}{ |p{3.6cm}|p{1.55cm}|p{1.55cm}|p{1.55cm}|p{1.55cm}|p{1.55cm}|}
\hline
\centering Models &  Accuracy & AUROC & Precision & Recall  & F1-score  \\
\hline
Random Forest   & 96.51  \%&  0.985  &   0.86   & 0.85 & 0.86  \\\hline
XGBoost         & 97.14  \%& 0.991   &   1.00   & 0.77 & 0.87  \\\hline
DNN             & 97.94  \%& 0.988   &   0.98   & 0.85 & 0.91  \\\hline
Isolation Forest& 89.99  \%&  0.680  &   0.65   & 0.39 & 0.49  \\\hline
Autoencoder     & 86.77  \%&  0.834  &   0.47   & 0.64 & 0.54  \\\hline
Capsule Network & \textbf{99.28} \% &  \textbf{0.999}  &   1.00   & 0.94 & \textbf{0.97}  \\
\hline
\end{tabular}
\end{center}
\caption{Test set results for Dataset-4}
\label{tab:st ens}
\end{table}

\begin{table}[H]
\centering
\begin{center}
\begin{tabular}{ |p{3.6cm}|p{1.55cm}|p{1.55cm}|p{1.55cm}|p{1.55cm}|p{1.55cm}|}
\hline
\centering Models &  Accuracy & AUROC & Precision & Recall  & F1-score  \\
\hline
Random Forest   & 97.37  \%& 0.995   &   0.96   & 0.84 & 0.89  \\\hline
XGBoost         & 95.33  \%& 0.993   &   0.99   & 0.66 & 0.79  \\\hline
DNN             & 98.25  \%&  0.997  &   0.98   & 0.89 & 0.93  \\\hline
Isolation Forest& 86.49  \%&  0.554  &   0.47   & 0.13 & 0.21  \\\hline
Autoencoder     & 80.17  \%&  0.827  &   0.37   & 0.73 & 0.50  \\\hline
Capsule Network & \textbf{99.32} \% & \textbf{ 0.999}  &   0.99   & 0.96 & \textbf{0.97 } \\
\hline
\end{tabular}
\end{center}
\caption{Test set results for Dataset-5}
\label{tab:st ens}
\end{table}

\begin{table}[H]
\centering
\begin{center}
\begin{tabular}{ |p{3.6cm}|p{1.55cm}|p{1.55cm}|p{1.55cm}|p{1.55cm}|p{1.55cm}|}
\hline
\centering Models &  Accuracy & AUROC & Precision & Recall  & F1-score  \\
\hline
Random Forest   & 97.90  \%&   0.996 &   0.91   & 0.92 & 0.92  \\\hline
XGBoost         & 98.08  \%&  0.997  &   0.95   & 0.89 & 0.92  \\\hline
DNN             & 98.36  \%&  0.997  &   0.96   & 0.91 & 0.93  \\\hline
Isolation Forest& 87.76  \%&  0.633  &   0.52   & 0.31 & 0.39  \\\hline
Autoencoder     & 87.76  \%&  0.877  &   0.51   & 0.68 & 0.58  \\\hline
Capsule Network & \textbf{99.18} \% &  \textbf{0.999}  &   0.98   & 0.96 & \textbf{0.97}  \\
\hline
\end{tabular}
\end{center}
\caption{Test set results for Dataset-6}
\label{tab:st ens}
\end{table}

\begin{table}[H]
\centering
\begin{center}
\begin{tabular}{ |p{3.6cm}|p{1.55cm}|p{1.55cm}|p{1.55cm}|p{1.55cm}|p{1.55cm}|}
\hline
\centering Models &  Accuracy & AUROC & Precision & Recall  & F1-score  \\
\hline
Random Forest   & 95.19  \%&  0.989  &   0.73   & 0.95 & 0.82  \\\hline
XGBoost         & 97.08  \%&  0.996  &   0.96   & 0.79 & 0.86  \\\hline
DNN             & 98.02  \%&  0.996  &   0.96   & 0.87 & 0.91  \\\hline
Isolation Forest& 92.27  \%&  0.766  &   0.72   & 0.56 & 0.63  \\\hline
Autoencoder     & 93.22  \%&  0.923  &   0.72   & 0.70 & 0.71  \\\hline
Capsule Network & \textbf{98.88} \% &   \textbf{0.999} &   0.98   & 0.92 & \textbf{0.95} \\
\hline
\end{tabular}
\end{center}
\caption{Test set results for Dataset-7}
\label{tab:st ens}
\end{table}

\begin{table}[H]
\centering
\begin{center}
\begin{tabular}{ |p{3.6cm}|p{1.55cm}|p{1.55cm}|p{1.55cm}|p{1.55cm}|p{1.55cm}|}
\hline
\centering Models &  Accuracy & AUROC & Precision & Recall  & F1-score  \\
\hline
Random Forest   & 95.05  \%&  0.982  &   0.80   & 0.80 & 0.80  \\\hline
XGBoost         & 94.15  \%&  0.989  &   0.99   & 0.53 & 0.69  \\\hline
DNN             & 95.32  \%&  0.992  &   1.00   & 0.62 & 0.77  \\\hline
Isolation Forest& 87.22  \%&  0.582  &   0.46   & 0.20 & 0.28  \\\hline
Autoencoder     & 61.21  \%&  0.656  &   0.19   & 0.67 & 0.30  \\\hline
Capsule Network & \textbf{98.65} \% & \textbf{0.996}   &   1.00   & 0.89 & \textbf{0.94}  \\
\hline
\end{tabular}
\end{center}
\caption{Test set results for Dataset-8}
\label{tab:st ens}
\end{table}

\begin{table}[H]
\centering
\begin{center}
\begin{tabular}{ |p{3.6cm}|p{1.55cm}|p{1.55cm}|p{1.55cm}|p{1.55cm}|p{1.55cm}|}
\hline
\centering Models &  Accuracy & AUROC & Precision & Recall  & F1-score  \\
\hline
Random Forest   & 94.42  \%&  0.975  &   0.75   & 0.80 & 0.77  \\\hline
XGBoost         & 95.03  \%&  0.983  &   0.95   & 0.61 & 0.75  \\\hline
DNN             & 97.38  \%&  0.991  &   0.95   & 0.82 & 0.88  \\\hline
Isolation Forest& 89.44  \%&  0.675  &   0.59   & 0.39 & 0.47  \\\hline
Autoencoder     & 90.22  \%&  0.912  &   0.57   & 0.72 & 0.64  \\\hline
Capsule Network & \textbf{98.25} \% &  \textbf{0.998}  &   0.99   & 0.86 & \textbf{0.92}  \\
\hline
\end{tabular}
\end{center}
\caption{Test set results for Dataset-9}
\label{tab:st ens}
\end{table}

\section{Second Appendix}\label{apd:second}

Here are the plots of the results obtained on test data of different models used here:

\begin{figure}
    \centering
    \includegraphics[scale=0.48]{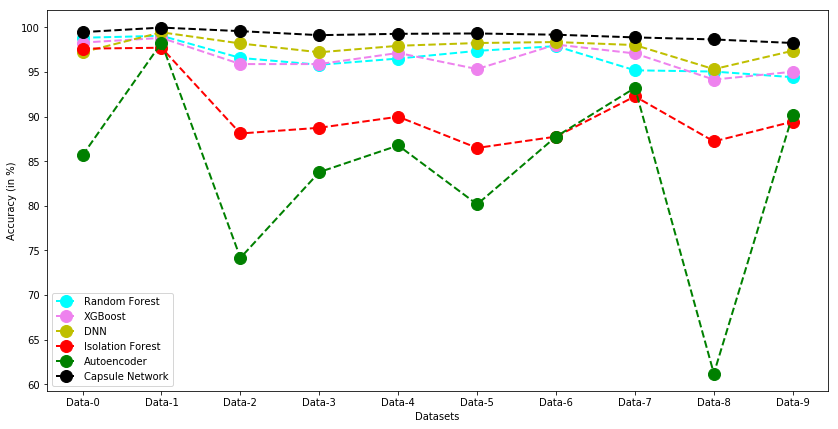}
    \caption{Accuracies of Capsule Network and baseline models on test datasets}
    \label{fig:accuracy}
\end{figure}

\end{document}